%% file: bmvc_final.tex
\documentclass{bmvc2k}

\usepackage{multirow}
\usepackage{amsfonts}

\title{DisCoM-KD: Cross-Modal Knowledge Distillation via Disentanglement Representation and Adversarial Learning}

\addauthor{Dino Ienco}{dino.ienco@inrae.fr}{1,2}
\addauthor{Cassio Fraga Dantas}{cassio.fraga-dantas@inrae.fr}{1,2}

\addinstitution{
 INRAE, UMR TETIS\\
 500, rue Jean Francois Breton\\
 34090, Montpellier, France
}
\addinstitution{
 INRIA, University of Montpellier\\
 860, rue Saint Priest\\
 34095, Montpellier, France
}

\runninghead{D. Ienco, C.F. Dantas}{Cross-Modal Knowledge Distillation}

\newcommand{\greenArrow}{ (\textcolor{green}{$\uparrow$}) }
\newcommand{\redArrow}{ (\textcolor{red}{$\downarrow$}) }
\newcommand{\method}{\textit{DisCoM-KD}}
\newcommand{\R}{\mathbb{R}}   
\newcommand{\dotp}[2]{\langle #1, #2 \rangle}

\begin{document}

\maketitle

\begin{abstract}
Cross-modal knowledge distillation (CMKD) refers to the scenario in which a learning framework must handle training and test data that exhibit a modality mismatch, more precisely, training and test data do not cover the same set of data modalities. Traditional approaches for CMKD are based on a teacher/student paradigm where a teacher is trained on multi-modal data with the aim to successively distill knowledge from a multi-modal teacher to a single-modal student. Despite the widespread adoption of such paradigm, recent research has highlighted its inherent limitations in the context of cross-modal knowledge transfer. \\
Taking a step beyond the teacher/student paradigm, here we introduce a new framework for cross-modal knowledge distillation, named \method{} (Disentanglement-learning based Cross-Modal Knowledge Distillation), that explicitly models different types of per-modality information with the aim to transfer knowledge from multi-modal data to a single-modal classifier. To this end, \method{} effectively combines disentanglement representation learning with adversarial domain adaptation to simultaneously extract, for each modality, domain-invariant, domain-informative and domain-irrelevant features according to a specific downstream task. Unlike the traditional teacher/student paradigm, our framework simultaneously learns all single-modal classifiers, eliminating the need to learn each student model separately as well as the teacher classifier.
We evaluated \method{} on three standard multi-modal benchmarks and compared its behaviour with recent SOTA knowledge distillation frameworks. The findings clearly demonstrate the effectiveness of \method{} over competitors considering mismatch scenarios involving both overlapping and non-overlapping modalities. These results offer insights to reconsider the traditional paradigm for distilling information from multi-modal data to single-modal neural networks. Our code is available at this \href{https://github.com/tanodino/CMKD_Disentangle}{link.} \end{abstract}

\section{Introduction}
\input{intro}

\section{Related Work}
\label{sec:related}
\input{related}

\section{Method}
\label{sec:method}
\input{method}

\section{Experimental Evaluation}
\label{sec:expe}
\input{expe}

\section{Conclusion}
\label{sec:conclu}
\input{conclu}

\section{Acknowledgment}
This work was supported by the French National Research Agency under the grant ANR-23-IAS1-0002 (ANR GEO ReSeT).

\bibliography{refs}
\end{document}

%% file: intro.tex
The modern landscape is characterized by a large diversity of devices consistently sensing their environments. This influx of information poses new challenges in terms of data analysis and understanding, particularly for machine learning and computer vision models, which must work seamlessly across a spectrum of platforms. From wearable gadgets to autonomous vehicles, an array of sensors continuously collect data about the surroundings~\cite{JabeenLSOLJ23}. 

In this context, the same object or entity can be described by multiple modalities, necessitating new learning paradigms to handle the collected heterogeneous information~\cite{LiangLFACMS23}. Many multi-modal learning models assume that data modalities between the training and deployment stages remain exactly the same~\cite{LiangLFACMS23}. However, due to the wide range of sensor data available, systematically accessing data across all sensor modalities may be infeasible.

Specifically, a set of modalities may be available during the training stage, while another set of modalities, either overlapping with the former or not, may be accessible during the deployment stage~\cite{McKinzieSCYST23}. In such a scenario, strategies are required to operate under cross-modal scenarios, leveraging the multi-modal information available during the training stage to enhance classification capabilities on the modalities accessible at deployment stage.

Recently, Cross-Modal Knowledge Distillation (CMKD) has proven to be effective for multi-modal applications characterized by modalities mismatch~\cite{LiuZZLQGDW23}. The majority of existing solutions rely on a teacher/student paradigm~\cite{SarkarE24,WangSKJZST24,JinHCM0Z23,HafnerBKG22}, where a teacher model trained on one or several data modalities is then used to supervise a single-modal student model trained on the modality available at deployment stage. However, this paradigm has several shortcomings, such as the arbitrary choice of modalities used to train the teacher model, the computational burden associated with the training of multiple models for specific downstream applications, and the need to set up a separate process each time a single-modal student model needs to be trained. Additionally, recent studies in~\cite{XueGRZ23} clearly highlights the inherent limitations of this paradigm in automatically distilling useful information such as modality-discriminative features for cross-modal knowledge transfer. 

In this study, we aim to address the challenge of multi-modal learning in applications characterized by modalities mismatch. We explore an alternative approach based on disentanglement representation and adversarial learning, overcoming mainstream teacher/student paradigms and their associated limitations. Specifically, we introduce a novel framework for cross-modal knowledge distillation, called \method{} (Disentanglement-learning based Cross-Modal Knowledge Distillation). This framework explicitly models different types of per-modality information to transfer knowledge from multi-modal data to a single-modal classifier. \method{} effectively combines disentanglement representation learning with adversarial domain adaptation to simultaneously extract domain-invariant, domain-informative, and domain-irrelevant features for each modality, tailored to a specific downstream task. Conversely to traditional teacher/student paradigms, our framework learns all single-modal classifiers simultaneously, eliminating the need to train each student model separately. To evaluate the effectiveness of our framework, we consider three standard multi-modal benchmarks and recent state-of-the-art knowledge distillation frameworks, demonstrating \method{}'s ability to outperform previous strategies in scenarios of modalities mismatch, covering both overlapping and non-overlapping modalities between training and deployment stages.  

In summary, the contributions of our work are the following: 
\begin{itemize}
\item A novel CMKD framework based on disentanglement representation and domain adversarial learning; 
\item An alternative CMKD strategy that circumvents traditional teacher/student paradigms to distill knowledge from multi-modal data to single-modal neural networks; 
\item An extensive comparison of \method{} across three computer vision multi-modal benchmarks with state-of-the-art knowledge distillation approaches, highlighting the broad applicability of our framework in the field of multi-modal learning.
\end{itemize}

The rest of the manuscript is organized as follows: related works are reviewed in Section~\ref{sec:related}. Section~\ref{sec:method} introduces the proposed framework based on feature disentanglement and adversarial learning. The experimental assessment and the related results are reported and discussed in Section~\ref{sec:expe}, while Section~\ref{sec:conclu} draws the conclusions.

%% file: related.tex
In this section, we firstly provide a brief recap of general knowledge distillation strategies, then we focus on knowledge distillation for multi-modal learning and, finally, we conclude by discussing elements on disentanglement representation-based learning.

\noindent \textbf{Knowledge Distillation.} Knowledge Distillation (KD)~\cite{HintonVD15} was introduced to transfer "dark" knowledge from a teacher model to a lightweight student model by learning the student model from the soft labels generated by the teacher. The standard KD loss formulation used to train the student model is defined as follows:
\begin{equation}
\label{eq:standard}
L = \alpha L_{task} + (1 - \alpha) L_{KD}
\end{equation}
Here, $L_{task}$ represents the downstream loss and $L_{KD}$ represents the distillation loss enforcing the knowledge transfer from teacher to student, with $\alpha$ determining the balance between the two terms. Different approaches vary in how they implement the $L_{KD}$ term, which can be logit-based~\cite{JinWL23}, feature-based~\cite{GuoYL023}, or relation-based~\cite{0020Y00022}. \\
Recent studies have shown that logit-based approaches outperform other strategies~\cite{abs-2403-01427}. Logit-based approaches implement the $L_{KD}$ component by exploiting the Kullback-Leibler divergence between the teacher and student logits. For example, \cite{ZhaoCSQL22} proposes to explicitly decouple the target class from non-target classes in the knowledge distillation process. In \cite{LiLYZSLLY23}, a curriculum learning process is introduced to estimate the temperature value to be used in the Kullback-Leibler divergence. \cite{JinWL23} employs a multi-level approach to perform logit distillation at different granularity levels (instance, class, and batch), considering multiple temperature values to make the knowledge transfer more robust. Recently, \cite{abs-2403-01427} proposes a plug-in extension that can be combined with any of the previous frameworks in which teacher and student logits are standardized prior to the analysis, ensuring a more coherent and consistent comparison. \\
\noindent \textbf{Cross/Multi-Modal Knowledge Distillation.} Cross-modal KD extends traditional KD approaches to encompass multi-modal learning~\cite{XueGRZ23}. While cross-modal KD does not assume any overlap between the modalities accessed by the teacher and student models, in the multi-modal KD scenario~\cite{LiuZZLQGDW23}, the information used by the student is a subset of the modalities used by the teacher model, thus making the former a more general scenario than the latter. However, in both scenarios, the student model typically has access only to a single data modality. The majority of the proposed frameworks for both cross-modal and multi-modal KD~\cite{SarkarE24,WangSKJZST24,JinHCM0Z23,HafnerBKG22} are tailored for task specific use cases with lack of a generic solution. This is primarily due to the fact that, despite the empirical success demonstrated by prior works, the mechanisms behind cross-modal KD still remains loosely understood~\cite{XueGRZ23}. An initial investigation towards understanding this mechanism has been proposed in~\cite{XueGRZ23}, where the authors emphasize the significance of modality-discriminative features as key components for cross-modal KD. However, their study provides preliminary experiments that are heavily reliant on data-specific characteristics, thus limiting the generic value of the obtained findings. \\
\noindent \textbf{Disentanglement Representation Learning.}
Disentangled representation learning aims to identify and separate hidden information in the underlying data~\cite{abs-2211-11695}. In contrast to standard learning processes, which focus solely on learning domain-invariant features, disentanglement-based methods explicitly decompose the learned representation into domain-specific and domain-invariant features, thus paving the way to the extraction of task-relevant and task-irrelevant information~\cite{Tenenbaum18}. 
In the context of multi-modal learning, disentanglement-based strategies are used to extract both multi-modal and modality-specific factors~\cite{TsaiLZMS19,ZhangZGCY23}. Recent approaches have focused on disentangling shared information among modalities to perform various downstream tasks~\cite{YuZWZLCX0M22,Materzynska0B22,XuLTLHSTGD22}. Despite its contributions to numerous settings~\cite{abs-2211-11695,JoY23,GhandehariounKL22}, disentangled representation learning still remains unexplored in the realm of knowledge distillation.

%% file: method.tex
\begin{figure}[!ht]
    \centering
    \includegraphics[width=1.\textwidth]{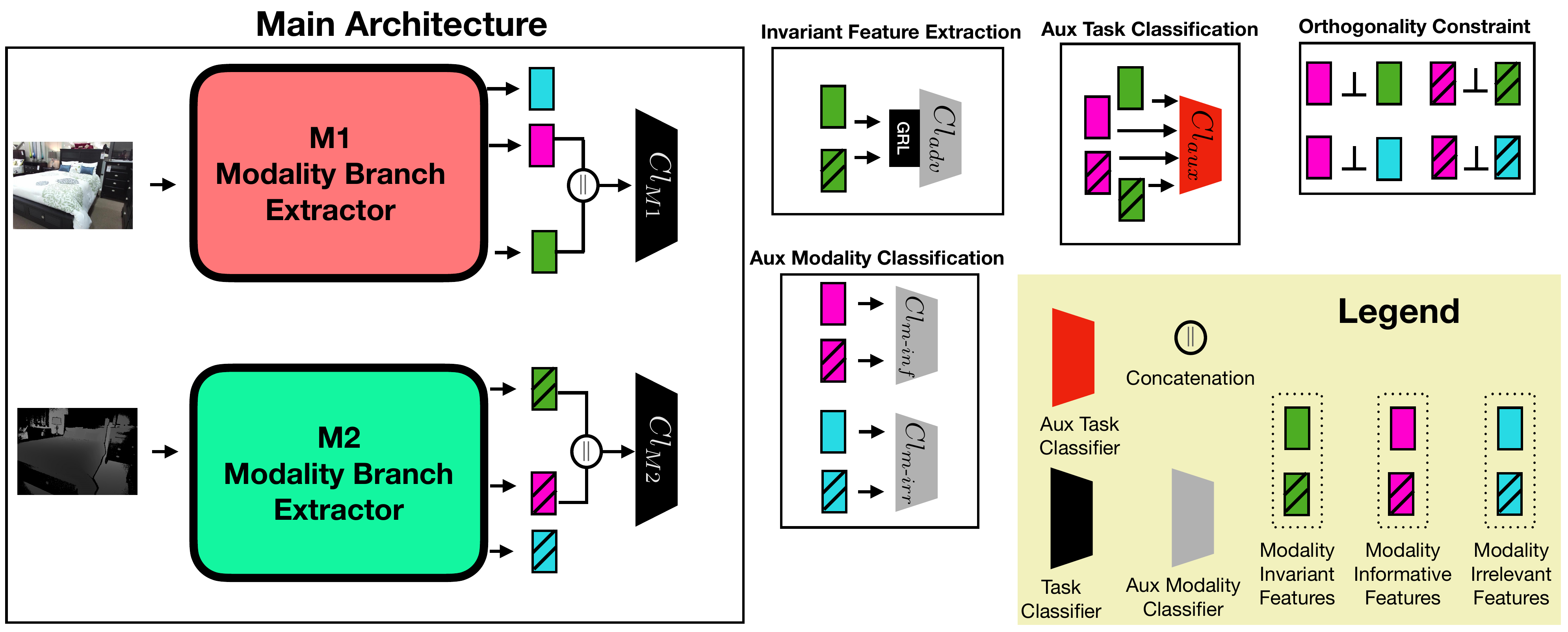}
    \caption{Schematic overview of \method{}: On the left, there are two per-modality branch extractors for modalities $M1$ and $M2$, along with two per-modality task classifiers to obtain the final prediction. On the right, several auxiliary classifiers, acting on intermediate representations, help disentangling per-modality information and make representations task informative. The training of the two parallel architectures is performed jointly, but at inference time, each model is deployed independently. \label{fig:archi} }
\end{figure}

The proposed architecture, depicted in Figure~\ref{fig:archi}, consists of two independent branches, one for each modality, extracting several per-modality representations. These representations are then used by per-modality task classifiers to make the final prediction. Furthermore, auxiliary classifiers, acting at intermediate stages, are leveraged to ensure that the extracted per-modality representations cover different complementary facets of the underlying information while also carrying information related to the downstream task.

Given an image $x_{*}$, where $x_{*}$ can be either $x_{M1}$ or $x_{M2}$  (with $M1$ and $M2$ being two different modalities), \method{} extracts three per-modality embeddings $z_{*}^{inv}, z_{*}^{inf}, z_{*}^{irr}$ referred as \textit{modality-invariant}, \textit{modality-informative} and \textit{modality-irrelevant} representation, respectively. All the embeddings have the same dimensionality $z_{*}^{inv}, z_{*}^{inf}, z_{*}^{irr} \in \R^{D}$. 
Subsequently, $z_{*}^{inv}$ and $z_{*}^{inf}$ are fed to the (per-modality) main task classifiers, while $z_{*}^{irr}$ is discarded, as its objective is to collect/attract per-modality information that should not contribute to the downstream task. Specifically, for each of the input modalities, we have a task classifier $Cl_{*}$ that outputs class probabilities $\hat{y}_{*} = Cl_{*}([z_{*}^{inv}||z_{*}^{inf}]) \in \R^{C}$ for the $C$ existing classes and $||$ denotes the concatenation operation. This means that two main task classifiers $Cl_{M1}$ and $Cl_{M2}$ are trained during the process, with $\hat{y}_{M1} =  Cl_{M1}([z_{M1}^{inv}||z_{M1}^{inf}])$ and $\hat{y}_{M2} = Cl_{M2}([z_{M2}^{inv}||z_{M2}^{inf}])$.

Beyond the main architecture, we introduce several modules to ensure that the extracted embeddings represent complementary information derived from the multi-modal input data. These additional modules are: i) A modality classifier $Cl_{adv}$ coupled with gradient reversal layer~\cite{ganin2016domain} to facilitate the extraction of modality-invariant representations; ii) Two auxiliary modality classifiers, $Cl_{m\text{-}inf}$ and $Cl_{m\text{-}irr}$, ensuring that modality-informative ($z_{*}^{inf}$) and modality-irrelevant ($z_{*}^{irr}$) embeddings contain modality-specific information and iii) An auxiliary task classifier $Cl_{aux}$ enforcing modality-invariant ($z_{*}^{inv}$) and modality-informative ($z_{*}^{inf}$) embeddings to be discriminative for the downstream task. During inference, only the per-modality extractors and the main task classifiers ($Cl_{M1}$ and $Cl_{M2}$) are retained, resulting in two distinct models that have been jointly learnt and can be deployed independently of each other.

\subsection{Training losses} \label{sec:losses}
To train our cross-modal knowledge distillation framework \method{}, we design a set of loss functions that explicitly model several properties beyond the main downstream classification task with the aim to enforce disentanglement across complementary per-modality representations. Specifically, the training procedure optimizes five different loss terms.

The first term is directly related to the downstream classification task. Both modality invariant ($z_{*}^{inv}$) and modality informative ($z_{*}^{inf}$) embeddings are fed to the main task classifiers. Then, we use standard Cross-Entropy loss (CE) between the output of each per-modality task classifier and the associated ground-truth $y$:
\begin{equation}
    \mathcal{L}_{cl} = \sum_{m \in \{M1,M2\} } CE(Cl_{m}([z_{m}^{inv}||z_{m}^{inf}]), y )
\end{equation}

The second term has the goal to enforce the learning of modality-invariant representations. We adopt a CE loss over the output of a classifier that discriminates across representations of samples from different modalities. Here, we use an adversarial training strategy, implemented via Gradient Reversal Layer (GRL)~\cite{ganin2016domain} over the modality invariant ($z_{*}^{inv}$) representations:
\begin{equation}
    \mathcal{L}_{adv} = \sum_{m \in \{M1,M2\} } CE(Cl_{adv}( GRL(z_{m}^{inv}) ), m )
\end{equation}

The third term guides the learning of modality-aware representations via modality classifiers. We use two classifiers, one for modality informative ($Cl_{m\text{-}inf}$) and one for modality irrelevant ($Cl_{m\text{-}irr}$) embeddings in order to predict from which branch the embedding originates:
\begin{equation}
    \mathcal{L}_{mod} = \sum_{m \in \{M1,M2\} } CE( Cl_{m\text{-}inf}(z_{m}^{inf}) , m ) + \sum_{m \in \{M1,M2\} } CE( Cl_{m\text{-}irr}(z_{m}^{irr}) , m )
\end{equation}

The fourth term aims to enhance the task discriminative information carried by the per-modality embeddings. Here, we employ an auxiliary task classifier over the set of modality invariant ($z_{*}^{inv}$) and modality informative ($z_{*}^{inf}$) embeddings:
\begin{equation}
    \mathcal{L}_{aux} = \sum_{m \in \{M1,M2\} } \sum_{i \in \{inv,inf\} } CE( Cl_{aux}(z_{m}^{i}) , y)
\end{equation}

The last term explicitly constrains embeddings from the same modality to contain complementary information. We implement a double disentanglement process, enforcing orthogonality~\cite{JoY23} between modality-invariant ($z_{*}^{inv}$) and informative ($z_{*}^{inf}$) representations, as well as between modality-informative ($z_{*}^{inf}$) and irrelevant ($z_{*}^{irr}$) embeddings. This guides the network to explicitly separate different per-modality contributions:
\begin{equation}
    \mathcal{L}_{\perp} = \sum_{m \in \{M1,M2\}} \frac{\dotp{z_{m}^{inv}}{z_{m}^{inf}}}{\|z_{m}^{inv}\|_2 \|z_{m}^{inf}\|_2} + \sum_{m \in \{M1,M2\}} \frac{\dotp{z_{m}^{inf}}{z_{m}^{irr}}}{\|z_{m}^{inf}\|_2 \|z_{m}^{irr}\|_2}
\end{equation}

The final loss function is defined as the sum of all the previous terms:
\begin{equation}
    \mathcal{L} = \mathcal{L}_{cl} + \mathcal{L}_{adv} + \mathcal{L}_{mod} + \mathcal{L}_{aux} + \mathcal{L}_{\perp}
\end{equation}

\begin{figure}[!ht]
\centering
    \includegraphics[width=1.\textwidth]{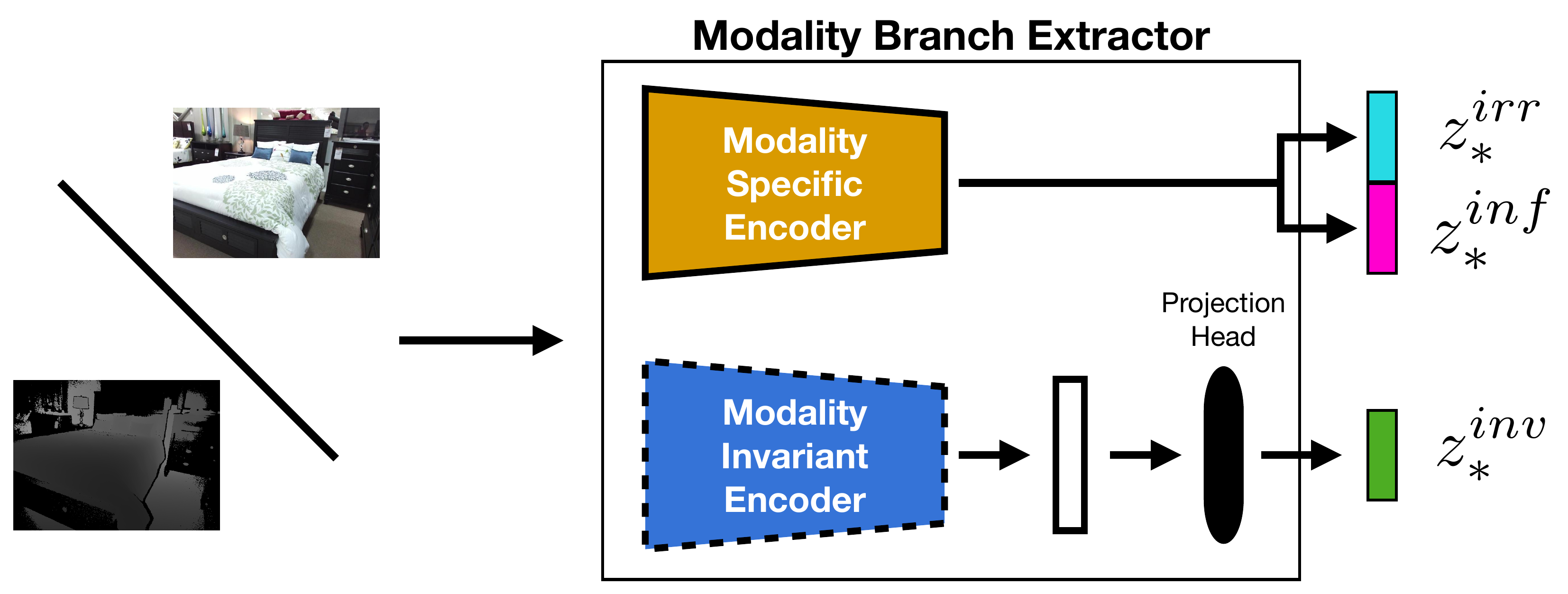}
    \caption{Details of the Modality Branch Extractor. It consists of two encoders, one extracting modality-specific ($z_{*}^{irr}$, $z_{*}^{inf}$) and one deriving modality-invariant ($z_{*}^{inv}$) representations. A projection head is used on the output of the modality-invariant encoder to obtain embeddings of the same size as the other representations.\label{fig:branch}}
\end{figure}

\noindent \textbf{Implementation details.} Each modality branch extractor, as reported in Figure~\ref{fig:branch}, is composed of two encoders: a modality specific encoder and a modality invariant encoder. As encoder backbone we use a ResNet-18 model~\cite{HeZRS16}.

The modality-specific encoder (shown in gold) extracts modality-specific information, namely: modality-irrelevant $z_{*}^{irr}$ and modality-informative $z_{*}^{inf}$ representations encoded separately into each half of the generated embedding vectors (depicted in light blue and purple, respectively). The modality-invariant encoder produces the modality-invariant representation $z_{*}^{inv}$. To ensure that all representations have the same size, a projection head, implemented via a fully connected layer, is used to project the output of the modality-invariant encoder to $z_{*}^{inv} \in \mathbb{R}^{D}$. For the two per-modality downstream task classifiers ($Cl_{M1}$ and $Cl_{M2}$) as well as all other auxiliary classifiers ($Cl_{adv}$, $Cl_{m\text{-}inf}$, $Cl_{m\text{-}irr}$, $Cl_{aux}$) we use a single linear layer with as many neurons as the number of classes to predict.

%% file: expe.tex
To evaluate our framework, \method{}, we designed an experimental evaluation considering three different multi-modal benchmarks involving SOTA teacher/student strategies from the Knowledge Distillation field encompassing both cross-modal and multi-modal KD scenarios. Additionally, an ablation study of \method{} is proposed to analyze the interplay between its different components.

\noindent \textbf{DATASETS.} As datasets we consider: 
i) \textbf{SUNRGBD}, the version proposed in~\cite{FerreriBT21} for multi-modal RGB-D scene classification. We consider RGB and Depth images from the Kinect v2 domain, for a total of 2\,105 pairs of RGB/Depth images, with 3 and 1 channels respectively, covering 10 classes; 
ii) \textbf{EuroSat-MS-SAR} proposed in~\cite{abs-2310-18653} for multi-modal Multi-Spectral (MS) and Synthetic Aperture Radar (SAR) remote sensing scene classification. The dataset contains 54\,000 pairs of MS and SAR images, with 13 and 2 channels respectively, for a land cover classification task spanning 10 classes;
iii) \textbf{TRISTAR} proposed in~\cite{StippelHK23} for multi-modal (RGB, Thermal and Depth) action recognition. According to results reported in~\cite{StippelHK23}, here we only consider the two most informative modalities (Thermal and Depth). The dataset contains 14\,201 pairs of Thermal and Depth images, with 1 channel each, representing an action recognition task spanning 6 classes.

\noindent \textbf{COMPETING METHODS.} We adopt three recent state-of-the-art strategies: Decoupled Knowledge distillation  ($DKD$)~\cite{ZhaoCSQL22}, Curriculum Temperature Knowledge Distillation ($CTKD$) \cite{LiLYZSLLY23} and Multi-Level Knowledge Distillation ($MLKD$)~\cite{JinWL23}. Furthermore, we integrate two baseline methods proposed in~\cite{XueGRZ23}, referred to as $KDv1$ and $KDv2$. Both baselines implement the traditional knowledge distillation loss reported in Equation~\ref{eq:standard}, with $KDv1$ setting the $\alpha$ hyper-parameter to 0, while $KDv2$ sets it to 0.5. While $KDv1$ only uses the soft label to train the student model, $KDv2$ equally weighs the information from the original hard labels and the teacher soft labels.
We combine each of these five strategies with the plug-in logit standardization preprocessing (LSKD) proposed in~\cite{abs-2403-01427}. Additionally, as references, we report the performance of the teacher model (referred to as TEACHER) and a student model that has not received any distillation supervision (referred to as STUDENT) for each evaluation scenario. 

\noindent \textbf{EVALUATION SCENARIOS}. We adopt two evaluation scenarios: cross-modal KD and multi-modal KD. For the cross-modal KD scenario the teacher is trained on the richest, in terms of downstream task performances, modality and, successively a single-modal student is distilled leveraging the remaining modality. Here the teacher is implemented via a ResNet-18~\cite{HeZRS16} architecture. For the multi-modal KD scenario the teacher model is trained on the full set of per-dataset modalities and, successively, a single-modal student is distilled. For this scenario, the teacher model is a two-branch architecture with a per modality encoder implemented via a ResNet-18. The fusion is performed at the penultimate layer of the ResNet-18 architecture via feature element-wise addition. Finally, a linear layer exploits the fused representation for the classification decision. All the student models are implemented with a ResNet-18 architecture.

\noindent \textbf{EXPERIMENTAL SETTINGS.}
For all the approaches the same training setup is used: 300 training epochs, a batch size of 128 and Adam~\cite{KingmaB14} as parameters optimizer with a learning rate of $10^{-4}$. For all the approaches we use online data augmentation via geometrical transformations (e.g. flipping and rotation). For the competing methods, we adopt the original hyper-parameter settings. The assessment of the models performance, on the test set, is done considering the weighted F1-Score, subsequently referred simply as F1-Score. Each dataset is divided into training, validation and test set with a proportion of 70\%, 10\% and 20\% of the original data, respectively. We repeat each experiment five times and report average results. Experiments are carried out on a workstation equipped with an Intel(R) Xeon(R) Gold 6226R CPU @ 2.90GHz, with 377Gb of RAM and four RTX3090 GPU. All the methods require only one GPU for training.

\subsection{Results}
Table~\ref{tab:crossmodal} and Table~\ref{tab:multimodal} present the average F1-Score results of the competing methods on both cross-modal and multi-modal KD scenarios, respectively. We use green arrows to indicate when a model outperforms the STUDENT baselines, and red arrows otherwise. \\
In the cross-modal KD scenario, Table~\ref{tab:crossmodal}, we consider the following cases: RGB $\rightarrow$ DEPTH for SUNRGBD, MS $\rightarrow$ SAR for EuroSat-MS-SAR, and THERMAL $\rightarrow$ DEPTH for TRISTAR. Here, the left modality indicates the one used by the teacher model while the one on the right is leveraged by the student. We observe that \method{} outperforms all competitors on both SUNRGBD (47.69 vs. 42.87 achieved by the best competitor) and EuroSat-MS-SAR (80.03 vs. 78.89 achieved by the best competitor). Moreover, it achieves comparable performance with the best competitor on TRISTAR (92.86 vs. 93.06 achieved by the best competitor). Notably, our framework is the only one that consistently improves (as indicated by green arrows) over the STUDENT baseline across all considered cross-modal scenarios.

In the multi-modal KD scenario, Table~\ref{tab:multimodal}, \method{} outperforms all state-of-the-art KD approaches, consistently improving classification performance compared to the STUDENT baseline. It is worth noting that our framework is the only one that achieves improvement on the TRISTAR dataset when the THERMAL modality is considered for the deployment stage, achieving a classification score of 97.06.
On EuroSat-MS-SAR, all competitors are capable of distilling a student single-modal neural network that outperforms the TEACHER model. Also in this case, \method{} achieves the best classification performance with a score of 98.12.
Interestingly, we observe that depending on the dataset, teachers trained on multiple modalities are not always the best choice for distilling a single-modal student. For example, in the TRISTAR case, when the deployment stage covers the DEPTH modality, all KD frameworks exhibited their best performances when the TEACHER has been only trained on the THERMAL modality (cross-modal KD scenario) rather than on the whole set of modalities (multi-modal KD scenario). This underscores that no strategy (neither cross-modal nor multi-modal) guarantees a systematic improvement, highlighting the arbitrary impact this inherent choice can have on the underlying distillation process.

\begin{table}[!ht]
    \centering
    \footnotesize
    \begin{tabular}{c|c||c|c||c|c||c|c|} \hline
       & & \textbf{SUNRGBD} & \textbf{EuroSat-MS-SAR} & \textbf{TRISTAR}\\ \hline
       & & RGB $\rightarrow$ DEPTH & MS $\rightarrow$ SAR &  THER. $\rightarrow$ DEPTH \\\hline \hline
      TEACHER & - & 44.45 & 95.49 & 96.99 \\ \hline
      STUDENT & - & 43.82 & 71.54 & 90.66\\ \hline \hline
    
      \multirow{2}{*}{KDv1} & ORIG & 40.10 \redArrow & 78.38 \greenArrow  & 92.71 \greenArrow \\ 
                             & w/ LSKD  & 37.96 \redArrow & 78.29 \greenArrow  & 91.91 \greenArrow \\ \hline
       \multirow{2}{*}{KDv2} & ORIG  & 41.61 \redArrow & 78.19 \greenArrow  & 92.92 \greenArrow \\ 
                             & w/ LSKD  & 42.08 \redArrow & 78.27 \greenArrow  & 91.68 \greenArrow \\ \hline
       \multirow{2}{*}{DKD} & ORIG  & 42.44 \redArrow & 78.30 \greenArrow  & 92.53 \greenArrow\\ 
                             & w/ LSKD  & 41.88 \redArrow & 78.83 \greenArrow  & 92.02 \greenArrow \\ \hline
       \multirow{2}{*}{CTKD} & ORIG  & 40.09 \redArrow & 78.89 \greenArrow  & 92.46 \greenArrow\\ 
                             & w/ LSKD  & 40.76 \redArrow & 78.89 \greenArrow  & 92.36 \greenArrow\\ \hline
       \multirow{2}{*}{MLKD} & ORIG  & 44.43 \greenArrow & 47.63 \redArrow & \textbf{93.06} \greenArrow\\ 
                             & w/ LSKD & 42.87 \redArrow & 78.13 \greenArrow  & 91.83 \greenArrow \\ \hline
        \method & - & \textbf{47.69} \greenArrow & \textbf{80.03} \greenArrow & 92.86 \greenArrow \\ \hline

    \end{tabular}
    \caption{Avg. F1-Score performances on cross-modal KD evaluation scenario. We consider the scenarios RGB $\rightarrow$ DEPTH, MS $\rightarrow$ SAR and THERMAL $\rightarrow$ DEPTH for the SUNRGBD, EuroSat-MS-SAR and TRISTAR, respectively. 
    \textcolor{green}{$\uparrow$} (resp. \textcolor{red}{$\downarrow$}
    ) indicates improved (resp. degraded) performances compared to the STUDENT baseline. \label{tab:crossmodal}}
    
\end{table}

\begin{table}[!ht]
    \centering
    \footnotesize
    \begin{tabular}{c|c||c|c||c|c||c|c|} \hline
       & & \multicolumn{2}{c||}{\textbf{SUNRGBD}} & \multicolumn{2}{c||}{\textbf{EuroSat-MS-SAR}} & \multicolumn{2}{c|}{\textbf{TRISTAR}}\\ \hline
       & & RGB & DEPTH & MS & SAR  & THER. & DEPTH\\ \hline \hline
      TEACHER & - & \multicolumn{2}{c||}{55.95}  & \multicolumn{2}{c||}{95.36 } & \multicolumn{2}{c|}{97.72 }\\ \hline
      STUDENT & - & 44.45 & 43.82 & 95.49 & 71.54  & 96.99 & 90.66 \\ \hline \hline
    
      \multirow{2}{*}{KDv1} & ORIG & 49.88 \greenArrow & 47.46 \greenArrow & 97.92 \greenArrow & 78.69 \greenArrow   & 96.82 \redArrow & 92.47 \greenArrow\\ 
                             & w/ LSKD  & 47.44 \greenArrow & 42.90 \redArrow & 97.37 \greenArrow & 78.45 \greenArrow  & 96.60 \redArrow & 91.45 \greenArrow\\ \hline
       \multirow{2}{*}{KDv2} & ORIG  & 50.38 \greenArrow & 46.08 \greenArrow & 97.90\greenArrow  & 78.86 \greenArrow  & 96.82 \redArrow & 92.54 \greenArrow\\ 
                             & w/ LSKD  & 47.38 \greenArrow & 43.52 \redArrow & 97.88 \greenArrow & 77.71 \greenArrow  & 96.22 \redArrow & 91.64 \greenArrow\\ \hline
       \multirow{2}{*}{DKD} & ORIG  & 48.95 \greenArrow & 46.38 \greenArrow & 97.39 \greenArrow & 78.45 \greenArrow  & 96.60 \redArrow & 91.54 \greenArrow\\ 
                             & w/ LSKD  & 49.01 \greenArrow & 43.40 \redArrow & 97.84 \greenArrow & 78.37 \greenArrow  & 96.54 \redArrow & 91.45 \greenArrow\\ \hline
       \multirow{2}{*}{CTKD} & ORIG  & 48.27 \greenArrow & 44.78 \greenArrow & 97.40 \greenArrow & 79.45 \greenArrow  & 91.84 \redArrow & 91.84 \greenArrow\\ 
                             & w/ LSKD  & 48.54\greenArrow  & 43.57 \redArrow  & 97.73 \greenArrow & 79.03 \greenArrow  & 96.57 \redArrow & 91.19 \greenArrow\\ \hline
       \multirow{2}{*}{MLKD} & ORIG  & 51.48 \greenArrow & 42.57 \redArrow  & 57.90 \redArrow  & 36.17 \redArrow   & 52.39 \redArrow & 92.27 \greenArrow\\ 
                             & w/ LSKD & 48.92 \greenArrow & 43.82 \redArrow  & 97.78 \greenArrow & 77.64 \greenArrow  & 91.64 \redArrow  & 91.44 \greenArrow \\ \hline
        \method & - & \textbf{53.63} \greenArrow & \textbf{47.69} \greenArrow & \textbf{98.12} \greenArrow & \textbf{80.03} \greenArrow  & \textbf{97.06} \greenArrow & \textbf{92.86} \greenArrow\\ \hline

    \end{tabular}
    \caption{ Avg. F1-Score performances on multi-modal KD evaluation scenario. Here, the TEACHER model has access to all modalities for each dataset.     \textcolor{green}{$\uparrow$} (resp. \textcolor{red}{$\downarrow$} ) indicates improved (resp. degraded) performances compared to the STUDENT baseline.\label{tab:multimodal} }
\end{table}

\noindent \textbf{Ablations.} 
The first ablation study (Table~\ref{tab:abla1}) explores the importance of the different components on which our framework is built. We observe that Auxiliary Task Classifiers ($\mathcal{L}_{aux}$) and the Disentanglement Loss ($\mathcal{L}_{\perp}$) seem to play the most significant roles in the underlying process. Depending on the considered dataset, each loss term has different relative impacts, and on average, the highest performance is achieved when all components are involved, underscoring the rationale behind the proposed framework. 
\begin{table}[!ht]
    \centering
    \small
    \begin{tabular}{l||c|c||c|c||c|c||c|} \hline
        & \multicolumn{2}{c||}{\textbf{SUNRGBD} } & \multicolumn{2}{c||}{\textbf{EuroSat} } & \multicolumn{2}{c||}{\textbf{TRISTAR} } & \textbf{Avg.} \\ \hline
        & RGB & DEPTH & MS & SAR & THER. & DEPTH & - \\ \hline
        w/o $\mathcal{L}_{adv}$ & 51.38 & 46.83 & 98.10 & 80.63 &  96.86 & 92.93 & 77.78\\ \hline
        w/o $\mathcal{L}_{mod}$ & 53.73 & 46.86 & 98.11 & 80.44 & 96.93 &  92.79 & 78.14\\ \hline
        w/o $\mathcal{L}_{aux}$ & 53.91 & 41.38& 97.82 & 79.10 &   96.68 & 91.56 & 76.74\\ \hline
        w/o $\mathcal{L}_{\perp}$ & 49.44 & 42.97 & 98.03 & 79.79 &   97.06 & 92.62 & 76.65\\ \hline \hline
        \method{} & 53.63 & 47.69 & 98.12 & 80.03 & 97.06 & 92.86 & \textbf{78.23}\\ \hline
    \end{tabular}
    \caption{\method{} components ablation study. 
    Analysis of the contributions of all the components on which our framework is built on in terms of Avg. F1-Score.\label{tab:abla1}}
\end{table}
The second ablation study (Table~\ref{tab:abla2}) investigates the interplay between the different representations extracted by the disentanglement process. Here, we note that only considering one of the two groups of information ---modality-invariant ($z_{*}^{inv}$) or modality-informative ($z_{*}^{inf}$)--- systematically decreases the classification performances. Modality-informative features provide slightly better discrimination capability, with varying margins depending on the dataset. In summary, this analysis suggests the suitability of exploiting both modality-invariant and modality-informative representations for the downstream classification task. 

\begin{table}[!ht]
    \centering
    \small
    \begin{tabular}{l||c|c||c|c||c|c||c|} \hline
        & \multicolumn{2}{c||}{\textbf{SUNRGBD} } & \multicolumn{2}{c||}{\textbf{EuroSat} } & \multicolumn{2}{c||}{\textbf{TRISTAR} } & \textbf{Avg.} \\ \hline
        & RGB & DEPTH & MS & SAR & THER. &  DEPTH & - \\ \hline
        Only $z_{*}^{inv}$ & 46.13 & 39.92 & 97.73 & 76.21 & 96.55 & 90.79 & 74.56 \\ \hline
        Only $z_{*}^{inf}$ & 50.82 & 43.29 & 97.60 & 78.09 & 96.20 & 90.68 & 76.11\\ \hline \hline
        \method{} & 53.63 & 47.69 & 98.12 & 80.03 &  97.06 & 92.86 & \textbf{78.23}\\ \hline
    \end{tabular}
    \caption{\method{} modality representations ablation study. Analysis of the contribution of the modality-invariant and -informative representations in terms of Avg. F1-Score. \label{tab:abla2}}
\end{table}

%% file: conclu.tex
In this study we have introduced a new framework for cross-modal knowledge distillation, namely \method{}. Our aim is to transfer knowledge from multi-modal data to a single-modal classifier.
To this end, our framework effectively combines disentanglement representation learning with adversarial domain adaptation.
Experimental evaluation, considering both cross-modal and multi-modal knowledge distillation evaluation scenarios, demonstrates the quality of \method{} compared to recent state-of-the-art KD techniques based on the standard teacher/student paradigm. In addition to performance improvements, our framework offers several inherent advantages over the standard paradigm: i) it learns all single-modal classifiers simultaneously, eliminating the need to train each student model separately; ii) it avoids the use of a teacher model, thereby eliminating the need to select which set of data modalities must be used to train the teacher model. Furthermore, our research work introduces an alternative strategy that opens new opportunities beyond the traditional teacher/student paradigm commonly employed for cross-modal and multi-modal knowledge distillation.

Several possible future avenues can be drawn. Our current process has only been assessed on cross-modal distillation tasks involving no more than two modalities. Extending \method{} to manage more than two modalities at once remains an open question. While most of the terms of the proposed loss function can be directly adapted to multiple modalities, how to modify the adversarial term to cope with more than two modalities is not straightforward. 
Another possible follow-up could investigate how to take inspiration from \method{} to design multi-modal distillation frameworks dealing with semantic segmentation and object detection tasks. For these tasks, the common methodologies are based on encoder/decoder neural network architectures that provide dense predictions as result. All these elements prevent the direct application of our methodology requiring to rethink how disentanglement and adversarial learning may be defined and implemented.